\documentclass{article}
\usepackage{spconf,amsmath,graphicx}
\usepackage{graphicx}
\usepackage{algorithm}  
\usepackage{algpseudocode}  
\usepackage{amsmath}  
\usepackage{cases}
\usepackage{indentfirst}
\usepackage{enumitem}
\usepackage{multirow}
\usepackage{gensymb}
\usepackage{hyperref}
\hypersetup{hypertex=true,
	colorlinks=true,
	linkcolor=green,
	anchorcolor=green,
	citecolor=green}

\title{Enhancing and Dissecting Crowd Counting By Synthetic Data}

\name{Yi Hou$ ^{1}$, Chengyang Li$ ^{2}$, Yuheng Lu$ ^{1}$, Liping Zhu$ ^{3}$, Yuan Li$ ^{1,\dagger}$, Huizhu Jia$ ^{1}$, Xiaodong Xie$ ^{1}$
	\thanks{$ ^{\dagger}  $This author is the corresponding author. }
}
\address{$ ^{1} $National Engineering Laboratory for Video Technology, Peking University\\
	$ ^{2} $School of Electronics Engineering and Computer Science, Peking University, Beijing, China\\
	$ ^{3} $Key Lab of Petroleum Data Mining, China University of Petroleum (Beijing)\\
	\{yihou, yuhenglu, yuanli, hzjia, donxie\}@pku.edu.cn, chengyang\_li@stu.pku.edu.cn, zhuliping@cup.edu.cn}

\begin{document}
	%
	\maketitle
	\begin{abstract}
		In this article, we propose a simulated crowd counting dataset CrowdX, which has a large scale, accurate labeling, parameterized realization, and high fidelity. The experimental results of using this dataset as data enhancement show that the performance of the proposed streamlined and efficient benchmark network ESA-Net can be improved by 8.4\%. The other two classic heterogeneous architectures MCNN and CSRNet pre-trained on CrowdX also show significant performance improvements. Considering many influencing factors determine performance, such as background, camera angle, human density, and resolution. Although these factors are important, there is still a lack of research on how they affect crowd counting. Thanks to the CrowdX dataset with rich annotation information, we conduct a large number of data-driven comparative experiments to analyze these factors. Our research provides a reference for a deeper understanding of the crowd counting problem and puts forward some useful suggestions in the actual deployment of the algorithm.
		
	\end{abstract}
	\section{Introduction}
	The application of crowd counting is very wide, from the construction of smart cities to the statistics of passenger flow in front of shops, all of which need the help of crowd counting algorithms.

	A lot of work has been proposed to improve the performance of detection algorithms. These studies either focus on proposing more advanced network structures (for example multi-column network \cite {zhang2016single, boominathan2016crowdnet}, scale aggregation module \cite {cao2018scale, zeng2017multi} and scale adaptive module \cite {liu2018geometric,wang2018defense,huang2018stacked}), or focus on designing more suitable loss functions \cite {ma2019bayesian}. These two focus points have greatly improved the performance of existing algorithms. In addition to the work on the algorithm side, on the dataset side, some open-source crowd counting datasets have also been proposed. The mainstream ones are UCFCC50 \cite {idrees2013multi}, ShanghaiTech A (ST\_A) \cite {zhang2016single}, UCSD \cite {chan2008privacy}, ShanghaiTech B (ST\_B) \cite {zhang2016single}, and WorldExpo'10 \cite {zhang2016data}. Among them, the number of people in UCFCC50 \cite {idrees2013multi} and ShanghaiTech A (ST\_A) \cite {zhang2016single} is very crowded, and some even reach 3000 people in a single picture. These two datasets are collected from the Internet, and there are no unified scene and shooting characteristics. The other three datasets are captured from surveillance cameras with large differences in viewing angles, so the number of people is generally not crowded.

	\begin{figure}[tb]
		\centering
		\includegraphics[width=1\linewidth]{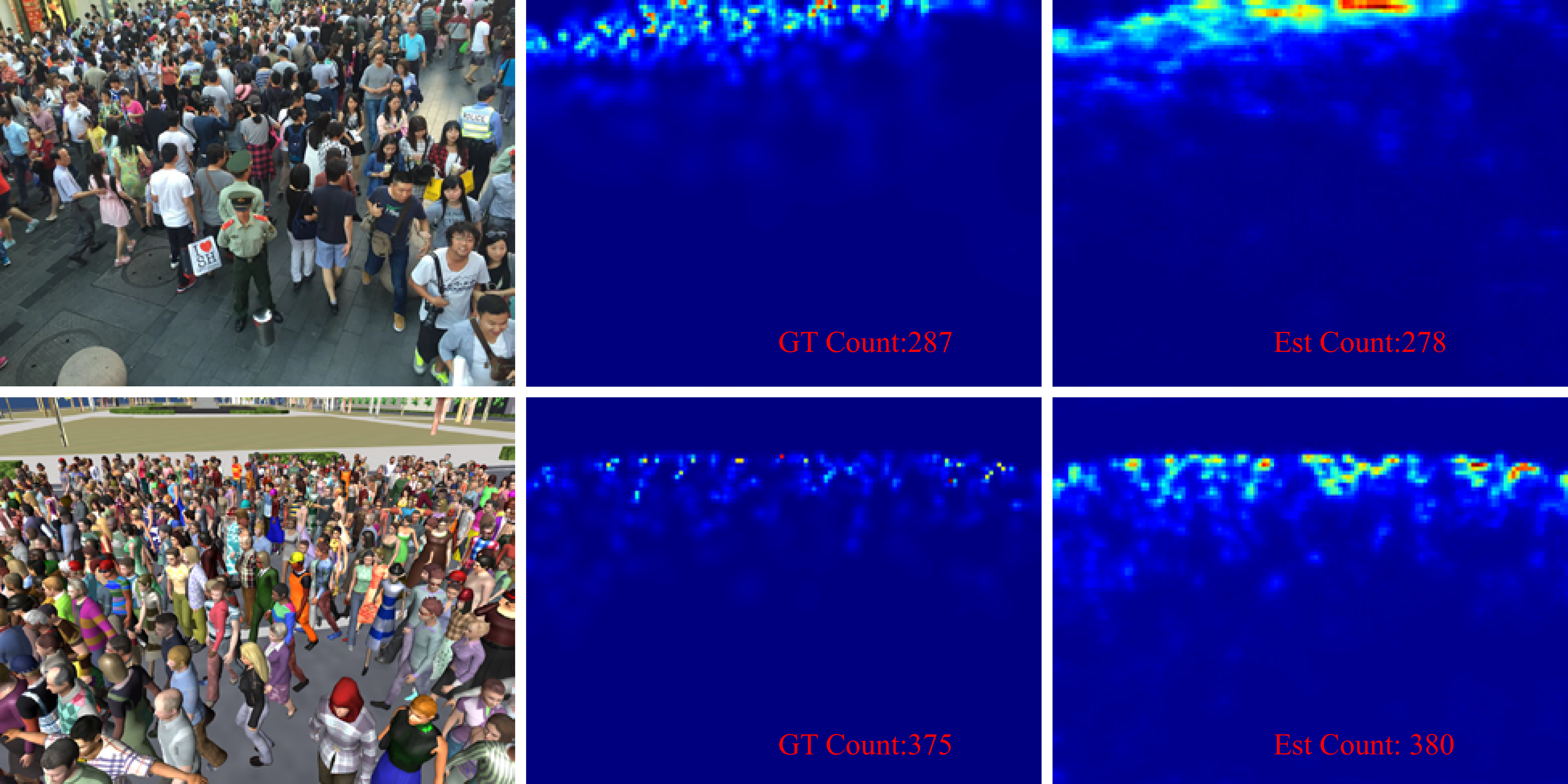}
		\caption{The left column shows a real image and a CrowdX image. The ground truth and the estimation results are shown in middle and right column.}
		\label{multi_factors}
	\end{figure}
	\begin{figure*}[tb]
		\centering
		\includegraphics[width=1\linewidth]{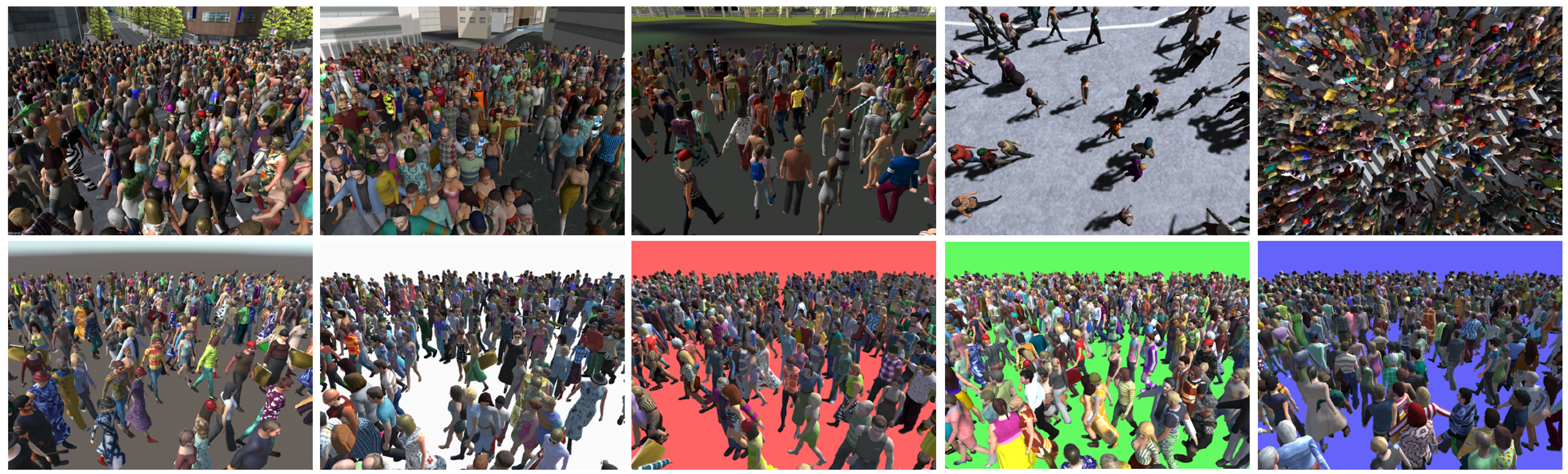}
		\caption{The proposed CrowdX dataset.}
		\label{main_show_pic}
	\end{figure*}

	These studies and datasets have improved the accuracy of crowd counting algorithms. However, these methods often have poor and unexplainable performance when they are migrated to the real environment. In summary, there are currently two issues that need to be resolved. 1) As shown in the table \ref {statistic_table}, the existing real-world datasets are generally not large, and insufficient training data will cause data-driven methods to be unable to learn diverse, complex, and complete features. To solve this problem, a simple method is to annotate large amounts of data. However, manually annotating them is very laborious and expensive. 2) Scene background, camera angle of view, and crowd density have a great influence on crowd counting algorithms. There is currently a lack of research on how they affect performance. For the first question, \cite {wang2019learning} proposed a simulation dataset GCC, which is constructed by game scenes and can be automatically labeled, but this method cannot control the key influence variables. For the second question, \cite {zhang2016single} proposed a method for head size adaptation, \cite {zhang2019multi} proposed an architecture for head resolution adaptation, and \cite {sam2017switching} proposes a training data selection strategy to improve algorithm performance. Although these methods have improved some performance, they only consider a single variable.
	
	In our work, we refer to the application of simulated datasets \cite{sankaranarayanan2018learning,gaidon2016virtual,li2018paralleleye,barbosa2018looking,ros2016synthia} in other fields, and construct a crowd simulation dataset that parameterizes the influencing factors to solve the first problem. To solve the second problem, we design a data-driven strategy to analyze the impact of key factors utilizing comparative experiments. Specifically, we propose a dataset generator based on the Unity3D\cite {unity_url}, which can parametrically control the influencing factors and generate a very realistic simulation dataset. We call the generated dataset CrowdX, which contains 24,000 automatically annotated crowd images. Compared with existing real-world datasets, CrowdX is easy to collect and annotate. Compared with the existing simulation dataset \cite {wang2019learning}, CrowdX is more realistic and can control influencing factors.
	
	In the experimental stage, to evaluate the effectiveness of using CrowdX for data enhancement, we first propose a streamlined and efficient benchmark network ESA-Net, then pre-train the network on CrowdX, and finally fine-tune the network with real-world datasets. The results show that the accuracy of the benchmark network on the ShanghaiTech B dataset has increased by 8.4\% and the accuracy has reached the state-of-the-art. In addition, the accuracy of the other two methods MCNN \cite {zhang2016single} and CSRNet \cite {li2018csrnet} has also been significantly improved, which verifies the effectiveness of using CrowdX as a data augmentation method. To evaluate the impact of the key factors in the performance, we have done many comparative experiments and find some interesting conclusions. Based on these findings, we provide some suggestions for selecting training datasets and deploying surveillance cameras.
	
	\begin{table}[]
		\centering
		\caption{Statistics of some widely used crowd datasets. PC means partially parameterizd and FC means fully parameterizd.}
		\scalebox{0.85}{
			\renewcommand\arraystretch{1.3}
			
			\begin{tabular}{l|c|c|c|c}
				\hline
				& Number & Resolution       & Average & Type    \\ \hline
				UCF\_CC\_50 \cite {idrees2013multi}     & 50     & 2101$\times$2888 & 1279    & Real    \\ \hline
				ST\_A \cite {zhang2016single}           & 482    & 589$\times$868   & 501     & Real    \\ \hline
				ST\_B \cite {zhang2016single}           & 716    & 768$\times$1024  & 123     & Real    \\ \hline
				WorldExpo \cite {zhang2016data}         & 3980   & 576$\times$720   & 50      & Real    \\ \hline
				UCSD\cite {chan2008privacy}            & 2000   & 158$\times$238   & 25      & Real    \\ \hline
				UCF\_QNRF \cite {idrees2018composition} & 1525   & 2013$\times$2902 & 815     & Real    \\ \hline
				GCC \cite {wang2019learning}            & 15212  & 1080$\times$1920 & 501     & Syn(PC) \\ \hline
				CrowdX                                                 & 24000  & 768$\times$1024  & 500     & Syn(FC) \\ \hline
			\end{tabular}
			
		}
		\label{statistic_table}
	\end{table}

	\begin{figure*}[tb]
		\centering
		\includegraphics[width=1\linewidth]{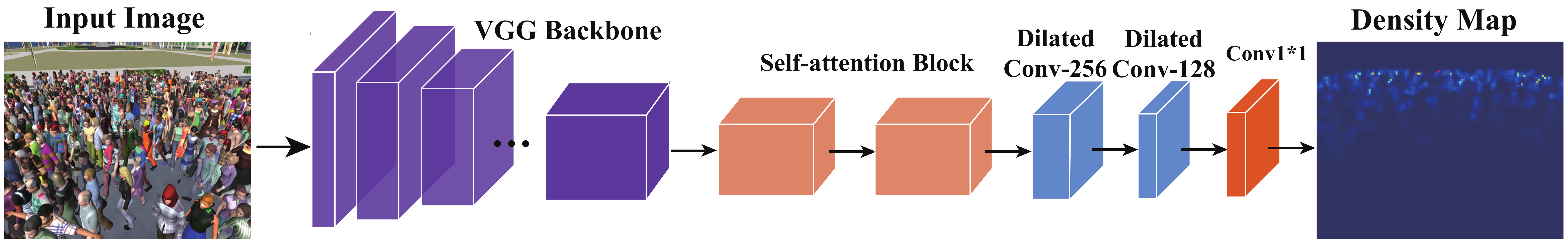}
		\caption{The proposed efficient self-attention network ESA-Net.}
		\label{network_table}
	\end{figure*}

	\section{Proposed Synsetic Dataset: CrowdX}
	\subsection{Assets and Tools Description}
	
	Based on the work of \cite{sun2019dissecting}, we choose the PersonX dataset originally used for person re-identification to construct CrowdX. PersonX contains 1,266 3D pedestrian models with diversified heights, weights, skin colors, hairstyles, and clothing. In addition to static properties, the dynamic features are also editable (such as standing, walking, and running), which makes the generated simulation dataset highly realistic. In terms of the simulation engine, we chose to use Unity3D to build the dataset. With its excellent rendering capabilities, we can simulate scenes, pedestrians, cameras, lights, etc. Considering that the crowd counting algorithm tends to be used in urban scenes, we select three urban scenes from the Unity3D asset store, all of which contain urban components such as buildings, urban roads, and traffic lights. In addition to these virtual backgrounds, we also use five solid color backgrounds to build CrowdX. The generated crowd image is shown in \ref {main_show_pic}.
	
	\subsection{Proposed Generation Method}
	For the virtual background, we manually select some sampling points from the three simulated cities to place pedestrians. These sampling points include pedestrian streets, parks, motor vehicles, etc., with strong diversity and high fidelity. For the solid color background, we directly sample 5 colors as the background. For each scene, we set the camera pitch angle to 30\degree, 50\degree, 70\degree, and 90\degree to take photos. To be consistent with the actual application and simplify the process, we set the roll value of the camera's Euler angle to 0 and fix the distance from the camera to the center of the scene. We sample a random number from 1 to 1000 and add the corresponding number of pedestrians to the scene with 100 as the step. The standing position is obtained by randomly sampling the visible area of the scene, and the standing direction is sampled according to a Gaussian distribution. The mean value of this Gaussian distribution is randomly generated, which can be considered as the main direction of the flow of people

	\subsection{Properties of CrowdX}
	As shown in the figure \ref {main_show_pic}, CrowdX is composed of crowd images with a simulated city background and a solid color background, with a total of 24,000 automatically labeled simulated images. The scale of this dataset exceeds other existing datasets and provides richer annotation information. Pedestrian annotation parameters include each pedestrian's id, 3D position, 2D camera plane position, height, and standing direction. The annotation parameters of the scene include the ID and type of the scene. The camera's annotation parameters include position, rotation, and rendering resolution.

	\section{Experiments}
	\subsection{Efficient benchmark Method: ESA-Net}
	
	The proposed ESA-Net architecture is shown in Fig \ref{network_table}, which consists of four parts, including 1) VGG16 \cite {SimonyanVery} based front-end for feature representation; 2) Two consecutive self-attention blocks for feature refining; 3) Two consecutive dilated convolutional layers used to expand the receptive field; 4) a $ 1 \times 1 $ convolutional layer at the end of the network, used to predict the density map. The network is streamlined and efficient enough, that is suitable to use in a production environment, compared with \cite{houyiIcassp2020}. We use L2 loss to train the benchmark network, and use mean absolute error (MAE) and mean square error (MSE) to evaluate performance, which is defined as:
	\begin{equation}\label{key}
		MAE =\frac{1}{N}\sum_{i=1}^{N}{\vert C_{i}-C_{i}^{GT} \vert} \ \ \ 
		MSE =\sqrt{\frac{1}{N}\sum_{i=1}^{N}{\vert C_{i}-C_{i}^{GT} \vert}^{2}}
	\end{equation}
	where $ C_{i} $ is the estiamted number of people, and $ C^{GT}_{i} $ is the groundtruth number. We pre-train the front-end on ImageNet \cite{deng2009imagenet}, and then train the benchmark network on the crowd datasets. We use a stochastic gradient descent (SGD) optimizer with a batch size of 50, set Nesterov momentum to 0.9, and set the weight decay to $ 5 \times 10 ^ {-4} $. The learning rate is initially set to $ 10 ^ {-6} $, and then divided by 10 every 3 epochs.

	\begin{table}[]
		\centering
		\caption{The effect of fine-tuning the models pre-trained on CrowdX on the real ShanghaiTech B dataset(MAE/MSE).}		
		\scalebox{0.9}{
			\begin{tabular}{c|c|c|c}
				\hline
				Method  & None/ImageNet & GCC       & CrowdX             \\ \hline
				MCNN    & 26.4/41.3     & 18.8/28.2 & \textbf{16.9/24.4}(\textcolor{red}{$ \downarrow $36.0\%})  \\ \hline
				CSRNet  & 10.6/16.0     & 10.1/15.7 & \textbf{9.7/14.6}(\textcolor{red}{$ \downarrow $8.5\%})   \\ \hline
				ESA-NET & 8.3/12.9      & -         & \textbf{7.6/11.8}(\textcolor{red}{$ \downarrow $8.4\%})   \\ \hline
		\end{tabular}}
		\label{finetune_table}
	\end{table}

	\subsection{Performence on CrowdX}
	
	We conduct two experiments to study the effectiveness of using CrowdX for data enhancement, including 1) training the benchmark model only on the ShanghaiTech B dataset, 2) pre-training on CrowdX, and then fine-tuning the model on the ShanghaiTech B dataset. The results show that the performance of the model pre-trained by CrowdX is improved from 8.3 to 7.6, and the accuracy is improved by about 8.4\%. We also test it on the ShanghaiTech B dataset, and the absolute error of the first 100 test samples shows that in almost all samples, the pre-trained model is significantly better than the non-pre-trained model, which further proves the effectiveness of the proposed CrowdX dataset in data enhancement.

	Furthermore, we conduct experiments on the other two latest models, MCNN \cite {zhang2016single} and CSRNet \cite {li2018csrnet}. Table \ref {finetune_table} shows that we achieve improvements on both architectures, with MCNN improved by 36\% and CSRNet by 8.5\%. Our method is better than GCC \cite {wang2019learning}, probably because the generation process of CrowdX has more comprehensive parameter control, and most of the background of CrowdX is an urban scene, which is consistent with the ShanghaiTech B.

	\begin{table}[]
		\centering
		\caption{The effect of the background scenes on experimental results. The value in row i and column j represents CMAE [dataset(i), dataset(j)], and the same below.}
		\scalebox{0.8}{
			\begin{tabular}{c|c|c|c}
				\hline
				& \begin{tabular}[c]{@{}c@{}}Solid Color Scene\end{tabular} & \begin{tabular}[c]{@{}c@{}}Synthetic Scene\end{tabular} & \begin{tabular}[c]{@{}c@{}}Real-world Scene\end{tabular} \\ \hline
				\begin{tabular}[c]{@{}c@{}}Solid Color\end{tabular} & \textbf{11.6(15.4)}                                               & { 61.2(109.8)}                              & 33.7(53.0)                                                       \\ \hline
				\begin{tabular}[c]{@{}c@{}}Synthetic\end{tabular}   & 33.3(41.8)                                                        & \textbf{14.9(23.1)}                                             & 37.8(68.6)                                                       \\ \hline
				\begin{tabular}[c]{@{}c@{}}Real-world\end{tabular}  & 75.7(91.6)                                                        & 158.0(200.6)                                                    & \textbf{15.9(23.1)}                                              \\ \hline
		\end{tabular}}

		\label{background}
	\end{table}
	
	\begin{table}[]
		\centering
		\caption{The effect of the camera perspective on experimental results. CP(30) is a dataset whose data is collected by controlling the camera perspective angle to equal 30\degree. CP(30,50) is the union set of CP(30) and CP(50). }
		\scalebox{0.6}{
			\renewcommand\arraystretch{1.3}
			\begin{tabular}{c|c|c|c|c|c|c}
				\hline
				& CP(30)       & CP(50)              & CP(70)              & CP(90)              & CP(30,50)    & ST\_B       \\ \hline
				CP(30)    & 20.1(28.4)   & \textbf{16.5(26.5)} & 38.4(47.9)          & 65.3(74.4)          & 18.3(27.5)   & 39.4(67.1)  \\ \hline
				CP(50)    & 41.7(60.7)   & \textbf{15.2(26.3)} & 21.6(26.8)          & 38.7(44.7)          & 39.3(67.7)   & 39.3(67.7)  \\ \hline
				CP(70)    & 134.5(167.3) & 52.1(69.0)          & 13.3(22.3)          & \textbf{12.6(18.0)} & 93.4(67.7)   & 57.2(94.0)  \\ \hline
				CP(90)    & 179.5(270.5) & 98.2(124.0)         & 21.7(36.0)          & \textbf{12.8(19.1)} & 138.9(178.0) & 67.8(107.3) \\ \hline
				CP(30,50) & 17.6(23.9)   & 12.3(22.4)          & \textbf{11.7(17.9)} & 29.2(35.3)          & 14.9(23.1)   & 37.8(68.6)  \\ \hline
			\end{tabular}
		}
		
		\label{cameral_perspective}
	\end{table}

	\subsection{Analysis of Influential Factors}
	In this section, we firstly define an operator CMAE $(\psi_{1},\psi_{2})$ to represent the result obtained by training the benchmark on the training set of $ \psi_ {1} $ and then evaluatingon the test set of $ \psi_ {2} $.
	
	\noindent
	\textbf{How does background affect performance?}
	From the diagonal of Table \ref {background}, we can know that the simpler the background of the dataset, the better the performance of the trained model. Besides, compared with the transfer from the simulated scene to the real scene, the benchmark model has better performance in the transfer from the solid color scene to the real scene, which indicates that the complex background hinders the feature transfer.

	\noindent
	\textbf{How does camera perspective affect performance?}
	Intuitively, the diagonal in Table \ref {cameral_perspective} should all be the best performance, but many exceptions can be found. Firstly, we find that CMAE [CP (30), CP (30)] $> $ CMAE [CP (30), CP (50)]. It can be inferred that the pitch value of the camera in CP (30) is too small, which increases the occlusion between people. The loss caused by occlusion is greater than the loss caused by the feature difference. In addition, we find that CMAE [CP(50), CP(ST B)] is smaller than CMAE [CP(70), CP(ST B)] and CMAE [CP(90), CP(ST B)]. We speculate that when the pitch value difference between the test set and the training set increases, the error will also increase. Furthermore, we find that the model trained on CP(30,50) is better than the model trained on a single CP(30) or CP(50). One possible reason is that although the camera pitch value is 30 degrees, there are still many people at a 50-degree angle.

	\noindent
	\textbf{How does person density affect performance?}
	We combine the data of CP (30) and CP (50) to form a dataset named PN (all). Then, we divide PN (all) into three sub-sets PN (0-200), PN (200-400), and PN (400+). PN (200-400) means that the number of people is between 200 and 400. The experimental results are shown in Table \ref {person_number}. We find that all models achieve the best performance on PN (0-200), which shows that the fewer people, the higher the accuracy. This is because the occlusion will be reduced when there are few people. In addition, we find that CMAE [PN(0-200), PN(400+)] is much larger than CMAE [PN(400+), PN(0-200)]. It seems that a model trained in a high-density scene is more versatile than a model trained in a low-density scene. We think this may be because the model trained on high-density datasets can learn occlusion features.

	\begin{table}[]
		\centering
		\caption{The effect of the number of people on experimental results. }
		\scalebox{0.8}{
			\begin{tabular}{c|c|c|c|c}
				\hline
				& \begin{tabular}[c]{@{}c@{}}PN(0-200)\end{tabular} & \begin{tabular}[c]{@{}c@{}}PN(200-400)\end{tabular} & \begin{tabular}[c]{@{}c@{}}PN(400+)\end{tabular} & \begin{tabular}[c]{@{}c@{}}PN(all)\end{tabular} \\ \hline
				\begin{tabular}[c]{@{}c@{}}PN(0-200)\end{tabular}   & \textbf{9.4(12.5)}                                   & {49.6(64.4)}                      & 154.8(173.6)                                        & 62.8(97.7)                                         \\ \hline
				\begin{tabular}[c]{@{}c@{}}PN(200-400)\end{tabular} & \textbf{13.4(17.7)}                                  & 17.6(22.1)                                             & 55.3(71.3)                                          & 26.6(41.8)                                         \\ \hline
				\begin{tabular}[c]{@{}c@{}}PN(400+)\end{tabular}    & \textbf{27.8(31.2)}                                  & 34.9(40.7)                                             & 33.2(44.9)                                          & 31.5(38.5)                                         \\ \hline
				\begin{tabular}[c]{@{}c@{}}PN(all)\end{tabular}     & \textbf{8.8(12.0)}                                   & 14.3(17.8)                                             & 23.2(34.9)                                          & 14.9(23.1)                                         \\ \hline
		\end{tabular}}

		\label{person_number}
	\end{table}
	\noindent
	\textbf{How dose image resolution affect counting performance?}
	
	We use two image resolutions ($ 1024\times 768 $) and ($ 512\times 384 $) to explore the effect of image resolution on the algorithm performance. To ignore the influence of factors such as background and perspective, we only use solid color background and camera perspective 30\degree in CrowdX. The results are shown in Table \ref{resolution}. It can be inferred from CMAE (Low resolution, Low resolution)$ < $CMAE (High resolution, High resolution) that large resolution is not necessarily helpful for improving the performance. From CMAE (Low resolution, High resolution) is much larger than CMAE (High resolution, Low resolution), we can see that models trained on low-resolution datasets are more versatile than models trained on high-resolution datasets.
	
	\begin{table}[]
		\centering
		\caption{The effect of image resolution on experimental results. }
		\scalebox{0.83}{
			\begin{tabular}{c|c|c}
				\hline
				& \begin{tabular}[c]{@{}c@{}}Low Resolution\end{tabular} & \begin{tabular}[c]{@{}c@{}}High Resolution\end{tabular} \\ \hline
				\begin{tabular}[c]{@{}c@{}}Low Resolution\end{tabular}  & \textbf{15.1(20.7)}                                      & {18.3(25.5)}                         \\ \hline
				\begin{tabular}[c]{@{}c@{}}High Resolution\end{tabular} & 32.7(43.2)                                               & \textbf{19.8(27.1)}                                       \\ \hline
		\end{tabular}}
		\label{resolution}
	\end{table}

	\section{Conclusion}
	
	In this article, we propose a synthetic crowd dataset CrowdX to enhance the level of crowd counting algorithms. With this dataset, the main influencing factors are analyzed and application suggestions are given. The conclusions of these experiments can improve the industry's understanding of the algorithm.

	\vfill\pagebreak
	
	\small
	\bibliographystyle{IEEEbib}
	\bibliography{strings,refs}
	
\end{document}